\documentclass[11pt]{article}

\usepackage{EACL2023}

\usepackage{times}
\usepackage{latexsym}
\usepackage{url}

\usepackage[T1]{fontenc}

\usepackage[utf8]{inputenc}

\usepackage{microtype}

\usepackage{inconsolata}
\usepackage{times}
\usepackage{latexsym}
\usepackage{graphicx}
\usepackage{multirow}
\usepackage{booktabs}

\title{Enhancing Dialogue Summarization with Topic-Aware Global- and Local- Level Centrality}

\author{Xinnian Liang\textsuperscript{1}\footnotemark[1], Shuangzhi Wu\textsuperscript{2}, Chenhao Cui\textsuperscript{2}, Jiaqi Bai\textsuperscript{1}, Chao Bian\textsuperscript{2}, Zhoujun Li\textsuperscript{1}\footnotemark[2]\\ 
\textsuperscript{1}State Key Lab of Software Development Environment, Beihang University, Beijing, China \\ 
\textsuperscript{2}Lark Platform Engineering-AI, Beijing, China\\ 
\texttt{\{xnliang,cuich,bjq,lzj\}@buaa.edu.cn,\{wufurui,zhangchaoyue.0\}@bytedance.com}
}

\begin{document} 
\maketitle 
\renewcommand{\thefootnote}{\fnsymbol{footnote}} 
\footnotetext[1]{Contribution during internship at ByteDance Inc.} 
\footnotetext[2]{Corresponding Authors.} 
\renewcommand{\thefootnote}{\arabic{footnote}} 

\begin{abstract}
Dialogue summarization aims to condense a given dialogue into a simple and focused summary text. Typically, both the roles' viewpoints and conversational topics change in the dialogue stream. Thus how to effectively handle the shifting topics and select the most salient utterance becomes one of the major challenges of this task. In this paper, we propose a novel topic-aware Global-Local Centrality (GLC) model to help select the salient context from all sub-topics. The centralities are constructed at both the global and local levels. The global one aims to identify vital sub-topics in the dialogue and the local one aims to select the most important context in each sub-topic. Specifically, the GLC collects sub-topic based on the utterance representations. And each utterance is aligned with one sub-topic. Based on the sub-topics, the GLC calculates global- and local-level centralities. Finally, we combine the two to guide the model to capture both salient context and sub-topics when generating summaries. Experimental results show that our model outperforms strong baselines on three public dialogue summarization datasets: CSDS, MC, and SAMSUM. Further analysis demonstrates that our GLC can exactly identify vital contents from sub-topics.~\footnote{\url{https://github.com/xnliang98/bart-glc}}
\end{abstract}

\section{Introduction}
\begin{figure*}
    \centering
    \includegraphics[width=\textwidth]{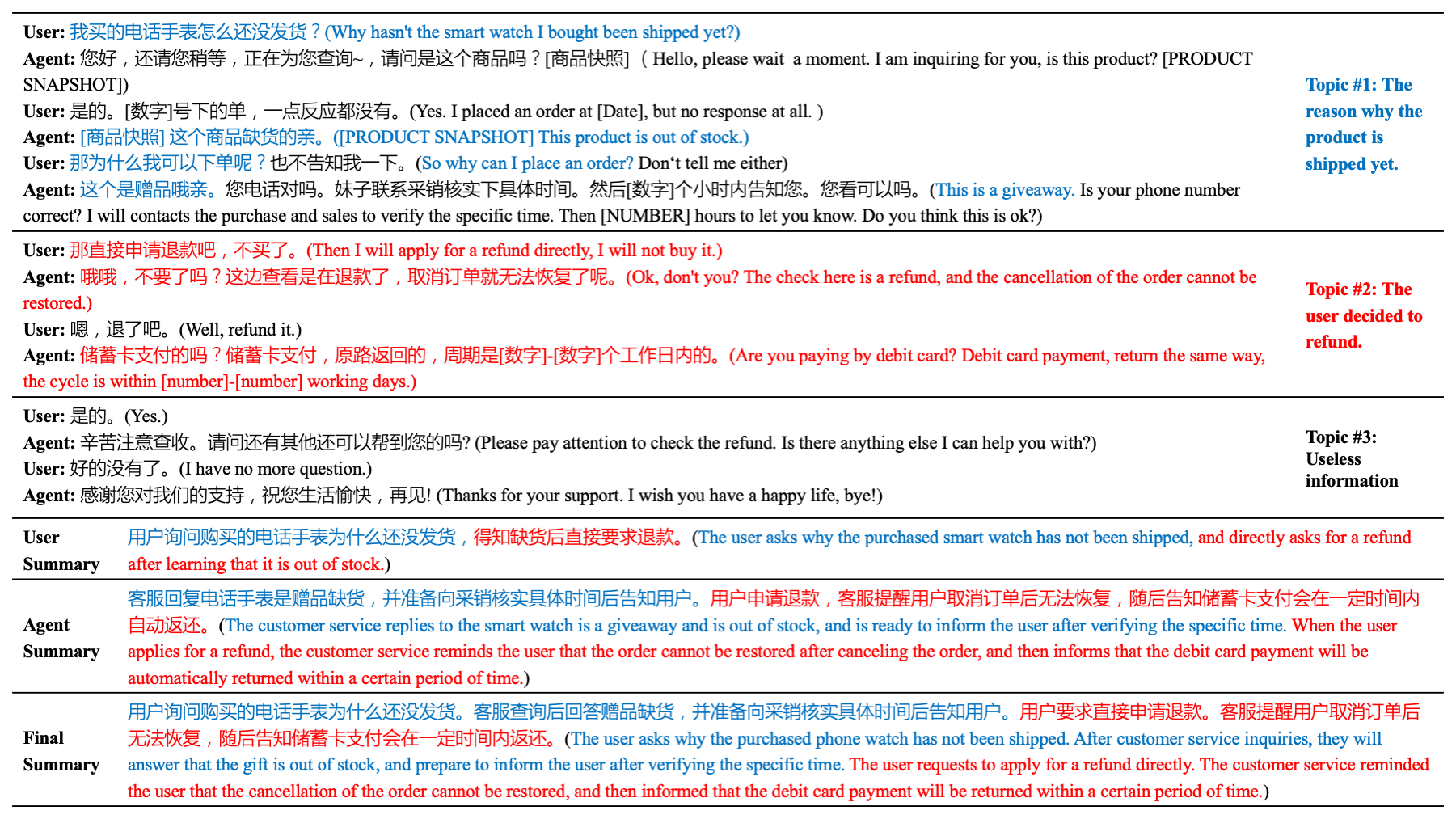}
    \caption{An example from the CSDS dataset. The dialogue contains 3 different sub-topics. The blue text represents sub-topic~\#1 and the red text represents sub-topic \#2. The sub-topic \#3 is useless information.}
    \label{fig:example}
\end{figure*}

Online conversations have become essential to communication in our daily work and life. Due to the information explosion, dialogue summarization has become a vivid field of research in recent years, which is meaningful for many applications, e.g. online customer service~\cite{10.1145/3292500.3330683,zhu-etal-2020-hierarchical} and meeting summary~\cite{feng-etal-2021-language}. 

Dialogue summarization aims to condense crucial information in a long dialogue into a short text like traditional summarization tasks. 
Differently, the main challenges of dialogue summarization are the viewpoints of multiple speaker roles~\cite{lin-etal-2021-csds,lin-etal-2022-roles,qi-etal-2021-improving-abstractive,zhang-etal-2022-summn} and shifting topics~\cite{chen-yang-2020-multi,Zou_Zhao_Kang_Lin_Peng_Jiang_Sun_Zhang_Huang_Liu_2021,liu-etal-2021-topic-aware} during the conversation process. 
As shown in Figure~\ref{fig:example}, summaries not only depend on the overall context but also needs the identification and selection of salient context in crucial sub-topics. We can see that the blue text in summaries is about sub-topic \#1 ``The reason why the product is shipped yet" and the orange text is about sub-topic \#2 ``The user decided to refund", which are aligned to the two salient sub-topics from dialogue utterances in the first and second block. The sub-topic \#3 is useless for summaries. This example shows the necessity to model the salient context and sub-topics in the dialogue. 

In this paper, we propose a novel topic-aware Global-Local Centrality (GLC) model to select salient contexts from all sub-topics. The centrality is an effective technique to measure the importance of sentences in a given document from unsupervised extractive summarization~\cite{zheng-lapata-2019-sentence,liang-etal-2021-improving,9664266}. 
The GLC contains global- and local-level centrality, which are used to capture the salience of sub-topics and content in each sub-topic respectively. Based on these centralities, we can guide the model to focus on the salient context and sub-topics when generating summaries.
Specifically, we employ utterance-level representations to cluster utterances and obtain sub-topic centers and assign each utterance to one sub-topic. 
Then, we compute the global centrality over sub-topic centers to measure the importance of each sub-topic and the local centrality over utterances of each sub-topic to measure the importance of sub-topic content. Finally, we combine the two to re-weight the dialogue context representations for the decoder to generate summaries. 

To evaluate the effectiveness of our proposed GLC, we apply the GLC to three different types of seq2seq structure: PGN, BERTAbs, and BART, and verify them on three public dialogue summarization datasets: CSDS, MC, and SAMSUM. 
CSDS and MC are two Chinese role-oriented summarization datasets that not only need to generate the overall summary of the dialogue but also need to generate role-oriented summaries for specific speakers in the dialogue as shown in Figure~\ref{fig:example}. 
SAMSUM is a widely used English dialogue summarization dataset. 
To generate role-oriented summaries, in this paper, we directly employ role prompts to guide the model to generate proper summaries. And the representations of role prompts can add role information to the centrality computation. 
Experimental results show that our GLC can improve the performance of all these seq2seq structures on three datasets. And the GLC-based BART model obtains new state-of-the-art results on the CSDS and MC.

Our contributions can be summarized as 1) We propose a novel topic-aware Global-Local Centrality (GLC) model to guide the model to identify the salient contexts and sub-topics in the dialogue. 2) Our GLC can bring improvement to different seq2seq models by easily plugging in and does not add any extra parameters to the seq2seq models. 3) The GLC-based BART model achieves new state-of-the-art results on CSDS and MC. Besides, extension studies prove our GLC can effectively capture vital sub-topics.

\section{Methodology}
\begin{figure*}
    \centering
    \includegraphics[width=0.85\textwidth]{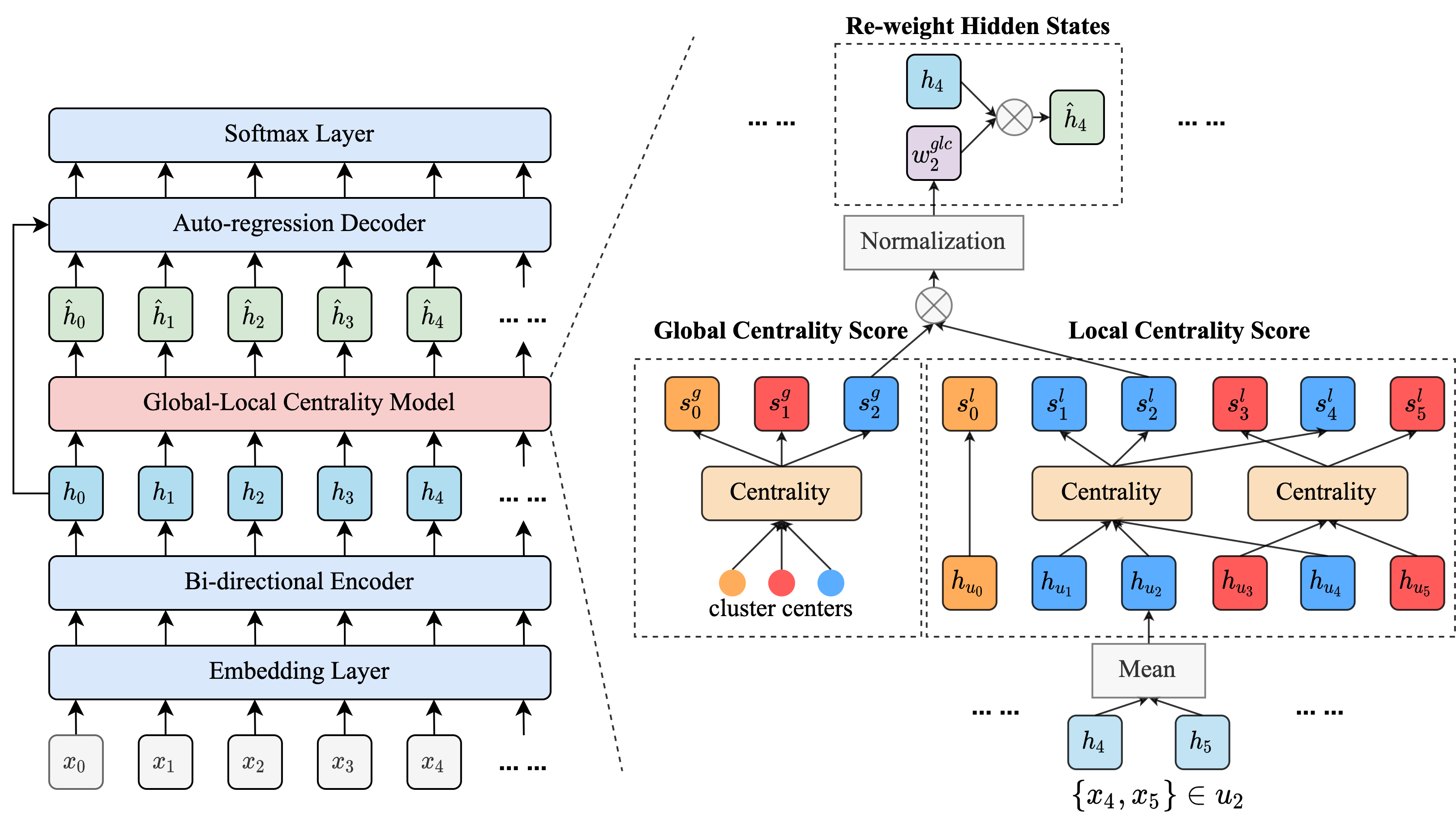}
    \caption{The main structure of our proposed method. The left is the framework of seq2seq with the GLC model. The right is the detailed process of our proposed GLC model.}
    \label{fig:main}
\end{figure*}
Figure~\ref{fig:main} shows the main structure of our proposed topic-aware global-local centrality (GLC) model. 
The seq2seq framework with GLC is on the left of Figure~\ref{fig:main}, which consists of the bi-directional encoder, global-local centrality model, and auto-regression decoder. 
The detail of GLC is on the right of Figure~\ref{fig:main}, which consists of global centrality and local centrality. In this section, we introduce them step by step.

\subsection{Task Formulation}
Firstly, we formulate the dialogue summarization task and role-oriented summarization task.
Given a dialogue $\mathcal D$ with $N$ utterances $\{u_1, \dots, u_N\}$ with $M$ roles $\{r_1,\dots,r_M\}$. Each utterance $u_i$ contains a speaker role $r_j$ and sentence $s_i$. We simply concatenate them by ``:'' and get utterance $u_i = r_j:s_i$. For role-oriented summarization tasks, the data contains different summary $y^{r_j}$ for different speaker roles $r_j$. In this paper, we employ $y^{user}$ and $y^{agent}$ to represent summaries of two different roles and $y^{final}$ to represent the overall summary of the whole dialogue. 
It is deserved to mention that our method can also be easily applied to datasets with more than two speaker roles by introducing different role prompts. 
Normal dialogue summarization task aims to generate overall summaries $y^{final}$ and role-oriented summarization task aims to generate role-specific summaries $y^{[user|agent|final]}$ from the input dialogue $\mathcal D = \{u_1, \dots, u_N\}$ according to the given role. 

\subsection{Role Prompts}
For role-oriented summarization tasks, previous works train multiple independent models for different role summaries, which is proven to hurt the performance of model~\cite{lin-etal-2022-roles} and needs more computation resources. 
In this paper, we employ a simple but effective trick to ensure that we only need to train a single model to obtain different role-specific summaries and overall summaries. Specifically, we use the prompts to control the generation of different kinds of summaries, which attach ``[User Summary]'', ``[Agent Summary]'', and ``[Final Summary]'' to the start of each dialogue as input to guide the model to generate required summaries. 
After that, the input context is re-formalized as ``[Role Prompt] Dialogue Contexts'' and then tokenized as $T$ tokens/words $\{x_t\}_{t=1}^T$ for the encoder of seq2seq model. 

\subsection{Bi-directional Encoder}
The bi-directional encoder is used to get tokens the semantic vector representations $\{h_t\}_{t=1}^T$ by capture bi-directional context information from tokens $\{x_t\}_{t=1}^T$ as follows:
\begin{equation}
    \{h_t\}_{t=1}^T = \mathtt{Encoder}(\{x_t\}_{t=1}^T)
\end{equation}
Then, we use the average of tokens vectors in each utterance as the semantic representations of dialogue utterances as follows:
\begin{equation}
    h_{u_i} = \frac{1}{|u_i|}\sum_t{x_t}, x_t \in u_i 
\end{equation}
After that, we can get the token-level semantic representations $\{h_t\}_{t=1}^T$ and the utterance-level semantic representations $\{h_{u_i}\}_{i=0}^{N}$, where $h_{u_0}$ is the vector representation of the attached role prompt, if role prompt is used.

\subsection{Global-Local Centrality Model}
Before feeding the representations into the decoder to generate the final summaries, we employ our proposed global-local centrality (GLC) model to re-weight the vector representations to identify salient facts in sub-topics over previous utterance-level semantic representations $\{h_{u_i}\}_{i=0}^{N}$. 

Firstly, our GLC obtains several cluster center points $\{c_k\}{k=1}^K$, which represent the center of sub-topics in the vector space. 
Then each utterance is assigned to the nearest center. 
As shown in Figure~\ref{fig:main}, utterances with the same color belong to the same sub-topic. We compute the global centrality score based on the cluster center representations to measure the importance of sub-topics and the local centrality score based on the utterance representations to measure the importance of each utterance belonging to the same sub-topic. 
Then, we employ their combination to get global-local centrality weights, which are used to re-weight the token-level vector representations. Finally, the re-weighted token-level vector representations are fed into the decoder to generate the summary. Our GLC can be directly plug-in any seq2seq structures, which makes it flexible.

\subsubsection{Obtain Cluster Centers}
To obtain the clusters, we directly call the K-Means algorithm, which is effective and widely used for cluster tasks.
And we all know setting the number of cluster centers for the K-Means algorithm is crucial and hard for the final results. 
However, we empirically find that we can set it as the number of utterances $(N+1)$ and then assign each utterance to the nearest cluster center point in the vector space.
After that, we find that many cluster centers have no assigned utterances and can be dropped. 
Based on this, we assume $K < (N+1)$ cluster centers $\{c_i\}_{i=1}^K$ are kept and note the vector representations of them as $\{h^c_i\}_{i=1}^K$.
\begin{equation}
    \{h^c_i\}_{i=1}^K = \mathtt{KMeans}(\{h_{u_i}\}_{i=0}^{N})
\end{equation}
And after the assignment of utterances, we can get $K$ clusters $\{\mathcal C_k\}_{i=k}^K$, which contain utterances with similar sub-topics. 
Each $\mathcal C_k$ contains several utterances and one cluster center point $c_k$. Through the previous method, we do not need to manually set the number of cluster centers for the K-Means algorithm.

\subsubsection{Global Centrality}
The global centrality score aims to measure the importance of each sub-topic by computing degree centrality based on the cluster center representations $\{h^c_k\}_{k=1}^K$. 
Each cluster center can be seen as one node on the graph, and the edge value between nodes $k$ and $j$ is $(h^c_k)^T \cdot h^c_j$. Then, the degree centrality of each cluster can be computed as follows:
\begin{equation}
    \mathtt{Cen}(c_k) = \sum_j (h^c_k)^T \cdot h^c_j
\end{equation}
where $\mathtt{Cen}(c_k)$ represents the importance of the cluster/sub-topic $k$ in the dialogue.
Then we normalize the score $\mathtt{Cen}(c_k)$ by $\frac{\mathtt{Cen}(c_k)}{||\{\mathtt{Cen}(c_k)\}_{k=1}^K||_2}$.

\subsubsection{Local Centrality}
The local centrality score aims to measure the importance of utterances in each cluster by computing the degree centrality. Each utterance can be seen as one node on the graph, and the edge value between nodes $i$ and $j$ is $(h_{u_i})^T \cdot h_{u_j}$. Then, the centrality of each utterance in the same cluster $\mathcal C_k$ can be computed as follows:
\begin{equation}
    \mathtt{Cen}(u_i) = \sum_j (h_{u_i})^T \cdot h_{u_j}, u_i, u_j \in \mathcal C_k
\end{equation}
where $\mathtt{Cen}(u_i)$ represents the importance of utterances in the $k$-th cluster/sub-topic. Then we normalize the score $\mathtt{Cen}(u_i)$ the same as the previous global centrality score.

\subsubsection{Global-Local Centrality Weight}
We can obtain the importance of each cluster (global centrality score) and the importance of utterances in each cluster (local centrality score) by the previous two steps. 
The most important utterance in the most important sub-topic should be assigned more attention when generating the summary.
So we obtain global-local centrality weight for each utterance in the dialogue by simply multiplying two centrality scores as follows:
\begin{equation}
w^{glc}_i = \mathtt{Cen}(u_i) \cdot \mathtt{Cen}(c_k), u_i \in \mathcal C_k
\end{equation}
Finally, we employ the global-local centrality weights to re-weight the token-level vector representations $\{h_t\}_{t=1}^T$ as follows:
\begin{equation}
    \hat h_t = w^{glc}_i \cdot h_t , x_t \in u_i
\end{equation}
Where each token uses the global-local centrality weight $w^{glc}_i$ of its utterance $u_i$ to re-weight the vector representation $h_t$. The token level representations $\{h_t\}_{t=1}^T$ are converted into $\{\hat h_t\}_{t=1}^T$.

\subsection{Auto-regression Decoder}
The auto-regression decoder generates the final summary based on the re-weighted context representations $\{\hat h_t\}_{t=1}^T$ as follows:
\begin{equation}
    P(\hat y) = \mathtt{Decoder}(\{\lambda\cdot\hat h_t + (1-\lambda)\cdot h_t\}_{t=1}^T)
\end{equation}where $\lambda$ is a hyper-parameter to control the influence of GLC, the default value of $\lambda$ is $0.5$.
In the training stage, the model learns the optimal parameters $\theta$ by minimizing the negative log-likelihood.

\section{Experiments}

\begin{table*}[]
\centering
\small

\begin{tabular}{@{}l|c|c|c|c|c@{}}
\toprule \midrule
\textbf{CSDS}      & \textbf{ROUGE-1}           & \textbf{ROUGE-2}           & \textbf{ROUGE-L}           & \textbf{BLEU}              & \textbf{BERTScore}         \\ \midrule
\textbf{PGN}       & 55.58/53.55/50.20          & 39.19/37.06/35.12          & 53.46/51.05/47.59          & 30.03/29.64/28.25          & 77.96/78.68/76.13          \\
\textbf{PGN-both}  & 57.20/56.08/51.62          & 40.37/39.10/36.50          & 55.14/53.85/49.12          & 32.58/33.54/29.78          & 78.69/79.52/76.74          \\
\textbf{PGN-GLC}   & 57.94/57.14/52.85          & 40.97/39.55/37.14          & 55.68/54.25/49.86          & 32.95/33.87.30.15          & 78.93/79.86/76.98          \\ \midrule
\textbf{BERT}      & 53.87/52.72/49.57          & 37.59/36.39/33.82          & 52.40/50.44/46.83          & 29.90/30.17/26.99          & 78.52/79.23/76.39          \\
\textbf{BERT-both} & 57.24/54.36/51.92          & 40.12/40.70/36.37          & 54.87/55.17/49.52          & 32.13/32.04/29.23          & 79.85/80.70/77.23          \\
\textbf{BERT-GLC}  & 57.59/55.14/52.34          & 41.28/41.84/36.48          & 55.74/55.86/50.16          & 32.75/32.64/29.81          & 79.89/80.71/77.28          \\ \midrule
\textbf{BART}      & 59.07/58.78/53.89          & 43.72/43.59/40.24          & 57.11/56.86/50.85          & 34.33/34.26/31.88          & 79.74/80.67/77.31          \\
\textbf{BART-both} & 59.21/58.93/54.01          & 43.88/43.69/\textbf{40.32}          & 57.32/57.28/51.10          & 34.75/34.49/32.30          & 79.72/80.64/77.30          \\
\textbf{BART-GLC}  & \textbf{60.07/61.42/54.59} & \textbf{44.67/45.83/}40.02 & \textbf{58.10/59.25/52.43} & \textbf{35.89/36.43/32.58} & \textbf{80.10/81.83/77.61} \\ \midrule \bottomrule
\end{tabular}
\caption{Results on the CSDS dataset test set. }
\label{tab:res_csds}
\end{table*}

\begin{table*}[ht]
\centering
\small
\begin{tabular}{@{}l|c|c|c|c|c@{}}
\toprule \midrule
\textbf{MC}        & \textbf{ROUGE-1}           & \textbf{ROUGE-2}           & \textbf{ROUGE-L}           & \textbf{BLEU}              & \textbf{BERTScore}         \\ \midrule
\textbf{PGN}       & 85.32/94.82/82.56          & 81.25/94.32/77.91          & 84.34/94.77/81.47          & 71.50/87.66/68.10          & 92.90/97.60/91.74          \\
\textbf{PGN-both}  & 85.98/95.10/83.37          & 81.93/94.59/78.78          & 84.94/95.06/82.20          & 72.77/87.82/69.63          & 93.23/97.71/92.15          \\
\textbf{PGN-GLC}   & 86.57/95.31/83.97          & 82.04/94.88/79.16          & 85.37/96.48/82.84          & 73.02/88.11/70.04          & 93.47/97.95/92.36          \\ \midrule
\textbf{BERT}      & 84.07/95.10/81.53          & 79.90/94.48/76.78          & 83.04/95.06/80.30          & 68.19/87.20/64.09          & 92.68/97.86/91.71          \\
\textbf{BERT-both} & 84.69/95.18/82.02          & 80.76/94.62/77.54          & 83.68/95.14/80.84          & 69.33/87.40/65.40          & 93.02/97.90/91.91          \\
\textbf{BERT-GLC}  & 85.64/95.49/82.87          & 81.44/94.97/78.05          & 84.16/96.10/81.57          & 69.84/87.94/66.01          & 93.15/97.92/92.36          \\ \midrule
\textbf{BART}      & 88.37/95.42/86.33          & 84.75/94.99/82.33          & 87.38/95.37/85.30          & 73.68/90.29/68.93          & 93.65/97.94/92.63          \\
\textbf{BART-both} & 88.52/95.63/87.06          & 85.22/95.42/82.89          & 87.75/95.91/85.78          & 73.87/90.70/69.31          & 93.69/97.88/92.69          \\
\textbf{BART-GLC}  & \textbf{89.55/96.84/88.47} & \textbf{86.47/96.14/84.62} & \textbf{88.56/96.23/86.77} & \textbf{74.19/91.32/70.18} & \textbf{94.17/98.25/92.96} \\ \midrule \bottomrule
\end{tabular}
\caption{Results on the MC dataset test set. }
\label{tab:res_mc}
\end{table*}

\begin{table*}[ht]
\centering
\begin{tabular}{@{}l|c|c|c|c|c@{}}
\toprule
\midrule
\textbf{SAMSUM}   & \textbf{ROUGE-1} & \textbf{ROUGE-2} & \textbf{ROUGE-L} & \textbf{BLEU}  & \textbf{BERTScore} \\ \midrule
\textbf{PGN}      & 40.08            & 15.28            & 36.63            & 37.49          & 80.67              \\
\textbf{PGN-GLC}  & 41.11            & 16.24            & 37.31            & 38.10           & 81.54              \\ \midrule
\textbf{BERT}     & 50.34            & 24.71            & 46.63            & 46.98          & 88.72              \\
\textbf{BERT-GLC} & 51.18            & 25.26            & 47.07            & 47.66          & 89.64              \\ \midrule
\textbf{BART}     & 53.12            & 27.95            & 49.15            & 49.28          & 92.14              \\
\textbf{BART-GLC} & \textbf{53.74}   & \textbf{28.83}   & \textbf{49.62}   & \textbf{50.36} & \textbf{92.77}     \\ 
\midrule
\bottomrule
\end{tabular}
\caption{Results on the SAMSUM dataset test set. }
\label{tab:res_samsum}
\end{table*}

\subsection{Datasets and Metrics}
We evaluate our method on three public datasets: CSDS~\cite{lin-etal-2021-csds}\footnote{https://github.com/xiaolinAndy/CSDS}, MC~\cite{song-etal-2020-summarizing}\footnote{https://github.com/cuhksz-nlp/HET-MC}, and SAMSUM~\cite{gliwa-etal-2019-samsum}\footnote{https://huggingface.co/datasets/samsum}.
The statistical information of them is shown in the appendix.
CSDS is the first role-oriented dialogue summarization dataset, which provides separate summaries for user and agent (customer service). MC is a Chinese medical inquiry dataset containing question summaries of patients and suggestion summaries of doctors. We note them as the user and agent summary.
For the MC dataset, we follow the data process and data split from RODS~\cite{lin-etal-2022-roles}. SAMSUM is a widely used English dialogue summarization dataset to evaluate the performance of models.

We employ lexical-level and semantic-level metrics to evaluate the performance of all models. Specifically, we use lexical level ROUGE-1/2/L~\cite{lin-2004-rouge}\footnote{https://pypi.org/project/rouge-score/} and BLEU~\cite{papineni-etal-2002-bleu}\footnote{https://github.com/mjpost/sacreBLEU}, which measure the similarity of references and generated summaries by computing the n-gram overlap of them. We use semantic level BERTScore~\cite{bert-score}\footnote{https://github.com/Tiiiger/bert\_score} and MoverScore~\cite{zhao2019moverscore}\footnote{https://github.com/AIPHES/emnlp19-moverscore}, which employ pre-trained language models to map the text into low-dimensional vectors in semantic space and then measure the similarity by computing the similarity by cosine similarity or word mover distance.
We can evaluate the performance of each model comprehensively through the previous metrics.
And all reported results are the average results of three different model checkpoints.
The results of Moverscore on three datasets can be found in the appendix.

\subsection{Baselines}
We applied our GLC on three widely used seq2seq models: PGN~\cite{see-etal-2017-get}, BERTAbs~\cite{liu-lapata-2019-text}, and BART~\cite{lewis-etal-2020-bart,shao2021cpt}.
\textbf{PGN} model is an LSTM-based seq2seq model without pre-training. \textbf{BERTAbs} is a BERT-based model, which employs BERT as the encoder and adds several transformer blocks as the decoder to generate summaries. We note it as BERT.
\textbf{BART} is a pre-trained transformer-based seq2seq model, which achieves the best results on many generation tasks. 
We add our proposed GLC into the previous three models and note them as \textbf{PGN-GLC}, \textbf{BERT-GLC}, and \textbf{BART-GLC}. 
We also compare our method with previous SOTA models: \textbf{PGN-both} and \textbf{BERT-both} from~\cite{lin-etal-2022-roles}, which proposed a role-interaction attention mechanism for the decoder. We reproduce it in the BART model as \textbf{BART-both}. For SAMSUM, we do not compare with BART-both due to this dataset does not contain role-oriented summaries.

\subsection{Implementation Details}
We use Chinese-BART-base\footnote{https://huggingface.co/uer/bart-base-chinese-cluecorpussmall} and BART-large\footnote{https://huggingface.co/facebook/bart-large} to initialize our transformer-based seq2seq model for Chinese and English datasets respectively.
We train all BART models on 4xV100 GPUs and PGN/BERT-based models on 1xV100 GPU.
For all models, the maximum input length is 512, the maximum generated summary length is 150, and the beam size is 3.
For BART-based models, the learning rate is 1e-4 with 10\% warmup steps, the total batch size is 64, and the training epochs are 5. For PGN/BERT-based models, we follow the settings from~\cite{lin-etal-2022-roles}. 

\subsection{Results}
The main results of the two role-oriented dialogue summarization datasets are shown in Table~\ref{tab:res_csds}-\ref{tab:res_mc}. Each block has three values, representing the final summary/user summary/agent summary from left to right. 
We can see that our proposed GLC can bring significant improvement to PGN, BERTAbs, and BART on the two datasets and BART-GLC achieves new state-of-the-art results. 
It is deserved to mention that our model does not need to modify any structure of the seq2seq structure and only needs to train one model for different summaries. We can see that the gain of metrics on the CSDS is better than on the MC, due to the summary of the MC dataset being highly similar to the input dialogue contexts. The results of the BERT-based model sometimes is worse than the PGN-based, we guess the reason is the prior knowledge learned in the pre-training stage of BERT is not suitable for the generation tasks. The improvement of lexical level metrics is more conspicuous than semantic level metrics due to the change of several words that may not affect the semantics of generated sentences. 
Overall, our proposed GLC is proved effective for the role-oriented dialogue summarization task with results on the two datasets.

The main results of the English dialogue summarization dataset are shown in Table~\ref{tab:res_samsum}. Because the SAMSUM does not provide role-specific summaries, we only report the performance of overall final summaries. From the results, we can see that our GLC can also bring significant improvements to three different seq2seq structures. 
We can see that the BERTScore is very high on SAUSUM, we guess that because the gold reference of this dataset is very short and this makes the semantic similarity between generated summaries and gold summaries close.
The results of SAMSUM demonstrated the effectiveness and generalization of our proposed method. 

\section{Discussion}
We conduct many external experiments on the CSDS dataset to further analyze the effectiveness of our proposed GLC. And more discussions are shown in the appendix.

\subsection{Ablation Study}
\begin{table}[]
\centering
\begin{tabular}{l|c}
\toprule \midrule
\textbf{}                            & \textbf{ROUGE-1}  \\ \midrule
\textbf{BART}                        & 59.07/58.78/53.89 \\ \midrule
\multicolumn{1}{l|}{\textbf{+Prompt}} & 59.42/58.96/54.03 \\
\multicolumn{1}{l|}{\textbf{+GC}}     & 59.64/59.55/54.24 \\
\multicolumn{1}{l|}{\textbf{+LC}}     & 59.37/59.47/54.11 \\
\multicolumn{1}{l|}{\textbf{+GC,LC}}     & 59.84/60.91/54.43 \\ \midrule
\textbf{BART-GLC}                    & \textbf{60.07/61.42/54.59} \\ \midrule \bottomrule
\end{tabular}
\caption{Ablation study on the CSDS dataset.}
\label{tab:ab}
\end{table}

To understand the impact of each component of our proposed GLC model, we compare the full BART-GLC with the following variants: (1) \textbf{BART:} three fine-tuned BART models for different summaries (final/user/agent); (2)\textbf{ BART+Prompt:} singe BART model with role prompts; (3) \textbf{BART+GC:} three BART models using global centrality scores to re-weight hidden states; (4) \textbf{BART+LC:} three BART models using local centrality scores to re-weight hidden states; (5) \textbf{BART+GC,LC:} three BART models using global-local centrality scores to re-weight hidden states. The results of these models are shown in Table~\ref{tab:ab}. From the results, we can see that all three components can bring improvement to the BART model, and the global-local centrality brings the greatest improvement. Interesting, The improvement brought by the combination of global and local centrality is far greater than the improvements they bring separately. This proves that global and local centrality are mutually beneficial.

\subsection{Human Evaluation}
\begin{table}[]
\centering
\begin{tabular}{@{}l|ccc@{}}
\toprule \midrule
                  & \textbf{Win} & \textbf{Loss} & \textbf{Tie} \\ \midrule
\textbf{CSDS\&MC} & 56.4         & 2.4           & 41.2         \\ \midrule
\textbf{SAMSUM}   & 51.8         & 3.2           & 45.3         \\ \midrule\bottomrule
\end{tabular}
\caption{Human evaluation results.}
\label{tab:human}
\end{table}
We use human evaluation~\cite{fang-etal-2022-spoken} to verify that our model outperforms the baseline. Specifically, we randomly sample 100 examples from three datasets and ask five NLP researchers to give a comparison between our model and baseline models. The evaluation results are represented as win, loss, and
tie, respectively indicating that the quality of generated summary by BART-GLC is better, weaker, or equal to the strong baselines.
Annotators were asked to judge from two aspects: fluency (whether contains grammatical and factual errors) and coverage (whether contains salient sub-topic information in the dialogue).
For two role-oriented dialogue summarization datasets CSDS and MC, our model is compared with BART-both. For SAMSUM, our model is compared with BART.
From the results in Table~\ref{tab:human}, we can see that our model is better than the baseline. Annotators tend to give ties on SAMSUM dataset. This may be caused by the length of summaries is short, which makes it hard to judge whether the summary is better or worse than the baseline model.

\subsection{Convergence of Training}
\begin{figure}
    \centering
    \includegraphics[width=0.45\textwidth]{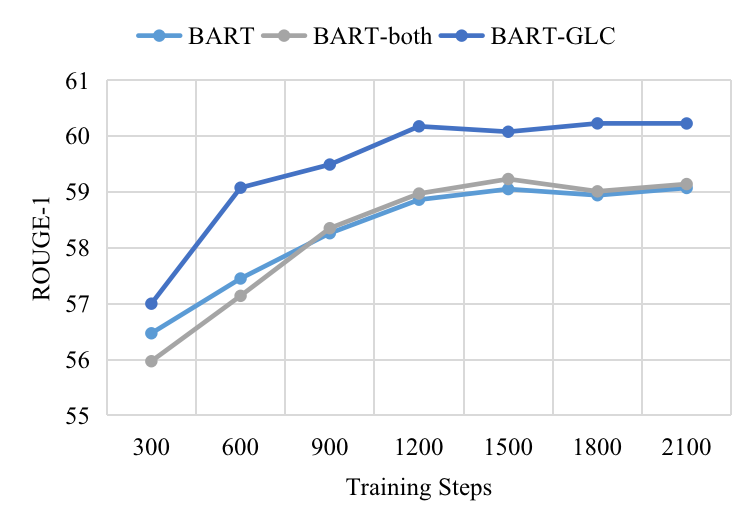}
    \caption{The ROUGE-1 score of different training step checkpoints.}
    \label{fig:training}
\end{figure}
We also compare the training convergence speed with BART and BART-both to prove our proposed GLC can bring effective prior knowledge for the seq2seq model. As shown in Figure~\ref{fig:training}, we can see that BART-GLC achieves comparable performance at 900 steps during training and reaches the SOTA results at 1,200 steps. This phenomenon demonstrates that our GLC brings prior knowledge into the model and speeds up the model training.

\subsection{Case study}
\begin{figure*}[ht]
    \centering
    \includegraphics[width=\textwidth]{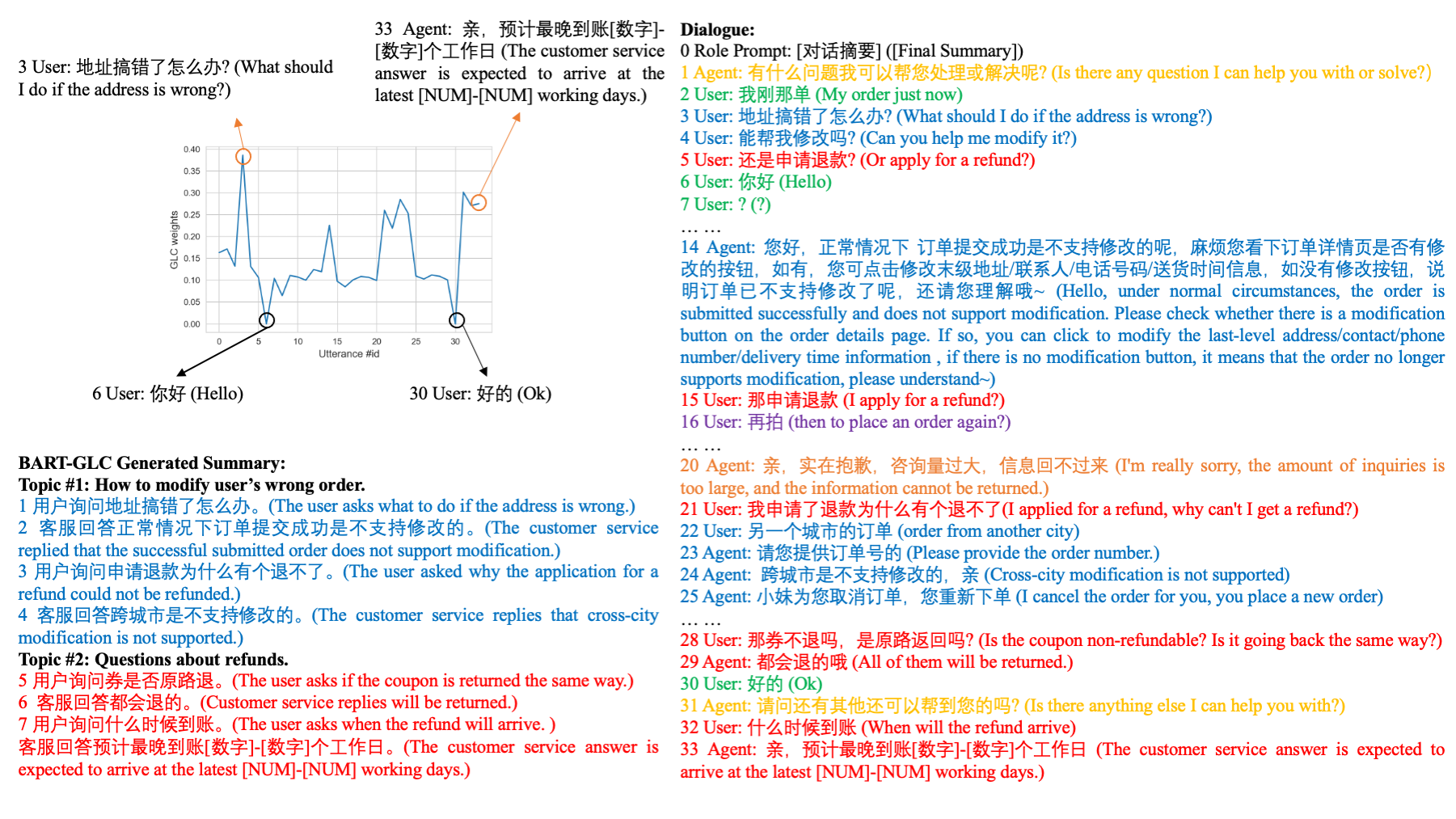}
    \caption{One case from the CSDS test set. Each color refers to one sub-topic. In the upper-left of this figure are the GLC weights and the corresponding utterances. In the bottom-left of the figure is generated summary of our proposed BART-GLC. On the right of the figure is the input dialogue.}
    \label{fig:case}
\end{figure*}
We select one example from the test set to show the ability of our proposed GLC in Figure~\ref{fig:case}. 
On the upper-left of this figure are the GLC weights and the corresponding utterances. In the bottom-left of the figure is generated summary of our proposed BART-GLC. On the right of the figure is the input dialogue and each color refers to one sub-topic. 
From this case, we can see that the final summary focus on two sub-topics: ``How to modify user's order'' and ``Questions about refunds''. And from the color on the right of this figure, we can see that our GLC can catch them accurately. Interestingly, generic utterances are aggregated into one topic (e.g. hello).
In the upper-left of this figure is the GLC weights and we can see that utterances, which are related to the final summary and belong to the vital sub-topics, are assigned high weights. This proves the global-local centrality exactly identified salient topics and utterances.

\section{Related Work}
Dialogue summarization has caught more and more attention in recent years and is widely used in various domains, e.g. meeting summarization~\cite{10.1007/11677482_3,feng-etal-2021-language}, daily dialogue summarization~\cite{krishna-etal-2021-generating,chen-etal-2021-dialogsum,zhong-etal-2021-qmsum}, etc.
Different from traditional summarization tasks, dialogue summarization needs to identify the role of speakers and capture the change of sub-topics during the dialogue. Besides, the dialogue summarization task has less labeled data and longer inputs. All of these make dialogue summarization harder to solve~\cite{chen-yang-2020-multi,zhang-etal-2021-exploratory-study,feng-etal-2021-language,lin-etal-2022-roles}. 

Recent dialogue summarization models can be categorized into three types: 1) data augmentation methods~\cite{feng-etal-2021-language,chen-yang-2021-simple,khalifa-etal-2021-bag}, which attempt to construct more pseudo-data to train a better model; 2) topic-based models \cite{Zou_Zhao_Kang_Lin_Peng_Jiang_Sun_Zhang_Huang_Liu_2021,liu-etal-2021-topic-aware,qi-etal-2021-improving-abstractive}, which track the change of topic information in the dialogue to generate more focused summary; and 3) semantic structure-based models \cite{liu-chen-2021-controllable,fu-etal-2021-repsum,Zhang_Zhang_Zaheer_Ahmed_2021,lei-etal-2021-finer-grain,zhao-etal-2021-give-truth,zhang-etal-2022-summn}, which employs semantic structures to enhance the summarization model. 

However, they ignored the sub-topics information in the dialogue utterances, which is crucial for dialogue summarization. Recently, \citet{zhao-etal-2020-improving} modified the attention mechanism to focus on the topic words, which can force the model to learn the topic information. \citet{Zou_Zhao_Kang_Lin_Peng_Jiang_Sun_Zhang_Huang_Liu_2021} employed Neural Topic Model to model the global level topic information. \citet{liu-etal-2021-topic-aware} tried to model the change of sub-topics by introducing contrastive learning.
Differently, in this paper, we bring the centrality, that has been widely used in unsupervised summarization~\cite{zheng-lapata-2019-sentence,liang-etal-2021-improving,9664266}, into the dialogue summarization task and proposed a novel topic-aware Global-Local Centrality model to capture salient dialogue utterances and sub-topics at the same time. Our proposed method is effective and more flexible.
 
\section{Conclusion}
In this paper, we bring the centrality into dialogue summarization tasks and proposed a novel topic-aware Global-Local Centrality (GLC) model for better capturing the sub-topic information in the dialogue utterances. Our GLC can be easily applied to any seq2seq structure and bring improvement to their performance. Experiments and further analysis demonstrated that GLC can effectively identify vital sub-topics and salient content in the dialogue. In future work, we will try to extend our work to datasets with longer inputs.

\section*{Limitations}
Our model also has some limitations: 1) The computation of sub-topic centers brings extra inference time into the basic seq2seq models. 2) We did not try to evaluate our model on longer dialogue summarization datasets. 3) We did not build a specific mechanism for different roles in role-oriented dialogue summarization task. We will try to solve these limitations in future work.

\section*{Acknowledgements}
This work was supported in part by the National Natural Science Foundation of China (Grant Nos. 62276017, U1636211, 61672081), the 2022 Tencent Big Travel Rhino-Bird Special Research Program, and the Fund of the State Key Laboratory of Software Development Environment (Grant No. SKLSDE-2021ZX-18).

\bibliography{anthology,custom}
\bibliographystyle{acl_natbib}

\appendix

\section{Datasets}
\begin{table}[ht]
\centering
\small
\begin{tabular}{l|c|c|c}
\toprule \midrule
                  & \textbf{CSDS}   & \textbf{MC}    & \textbf{SAMSUM} \\ \midrule
\textbf{Train Size}        & 9,101  & 29,324 & 14,732 \\
\textbf{Val. Size }         & 800    & 3,258 & 818 \\
\textbf{Test Size}         & 800    & 8,146 &  819\\
\textbf{Input Length}      & 321.92 & 292.21 & 94.52 \\
\textbf{User Sum. Length}  & 37.28  & 22.37  & - \\
\textbf{Agent Sum. Length} & 48.08  & 95.32  & - \\
\textbf{Final Sum. Length} & 83.21  & 114.54 &  20.34 \\ \midrule
\bottomrule
\end{tabular}
\caption{Statistical information of three datasets.}
\label{tab:datasets}
\end{table}
The statistical information of three datasets is shown in Table~\ref{tab:datasets}.

\section{Moverscore Results}

\begin{table*}[ht]
\centering
\begin{tabular}{@{}l|c|c|c@{}}
\toprule \midrule
\textbf{MoverScore} & \textbf{CSDS}              & \textbf{MC}                & \textbf{SAMSUM} \\ \midrule
\textbf{PGN}        & 59.00/58.68/58.23          & 80.90/93.84/79.69          & 59.87           \\
\textbf{PGN-both}   & 59.48/59.32/58.64          & 81.67/94.04/80.52          & -               \\
\textbf{PGN-GLC}    & 59.67/59.51/58.85          & 81.97/94.45/80.84          & 60.04           \\ \midrule
\textbf{BERT}       & 58.23/58.10/57.79          & 81.28/93.90/80.48          & 61.17           \\
\textbf{BERT-both}  & 59.52/59.55/58.46          & 82.26/94.20/81.02          & -               \\
\textbf{BERT-GLC}   & 59.74/59.62/58.90          & 82.64/94.49/81.44          & 61.59           \\ \midrule
\textbf{BART}       & 60.11/59.86/58.75          & 82.35/94.17/81.27          & 62.04           \\
\textbf{BART-both}  & 60.12/59.86/58.73          & 82.32/94.02/81.40          & -               \\
\textbf{BART-GLC}   & \textbf{60.32/61.03/59.02} & \textbf{82.94/95.35/82.10} & \textbf{62.27}  \\ \midrule \bottomrule
\end{tabular}

\caption{MoverScore on three datasets.}
\label{tab:moverscore}
\end{table*}
For Moverscore, we employ chinese-bert-wwm-ext\footnote{https://huggingface.co/hfl/chinese-bert-wwm-ext} to get the contextual embeddings of Chinese text input. Because \citet{lin-etal-2021-csds} did not provide they use what Chinese representation model, we use chinese-bert-wwm-ext to re-evaluate all their results and report in Table~\ref{tab:moverscore}.

\subsection{How abstractive is our model?}
\begin{figure}
    \centering
    \includegraphics[width=0.42\textwidth]{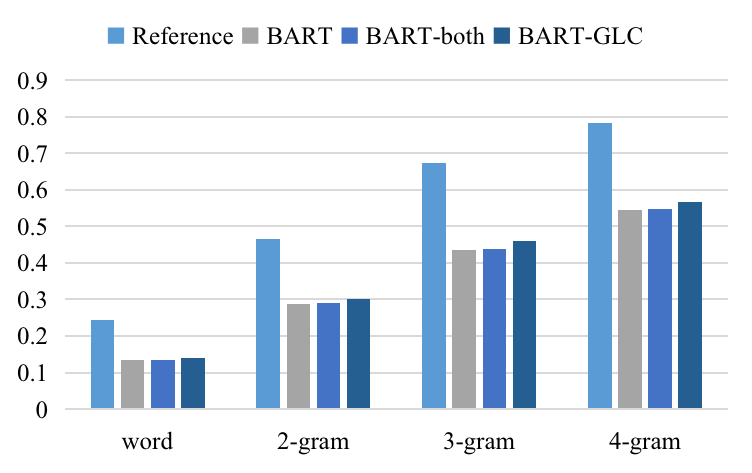}
    \caption{Percentage of novel words/n-grams in the reference and generated summaries of the CSDS test set.}
    \label{fig:novel}
\end{figure}
An abstractive model can be innovative by using words that are not from the input document in the summary. We measure the abstractive by the ratio of novel words or n-gram phrases in the summary. A higher ratio means a more abstractive model. We show the results in Figure~\ref{fig:novel}. We can see that our BART+GLC is more attractive than BART and BART-both. However, all of them have a big margin compared with references. It means more research is needed for generating more abstractive summaries.

\end{document}